\documentclass[11pt,a4paper]{article}
\usepackage[hyperref]{acl2021}
\usepackage{times}
\usepackage{latexsym}

\usepackage{microtype}

\aclfinalcopy 
\def\aclpaperid{3524} 

\usepackage{nicefrac} 
\usepackage{latexsym}
\usepackage{multirow}
\usepackage{booktabs}
\usepackage{amsmath}
\usepackage{booktabs}
\usepackage{subfig}
\usepackage{graphicx}
\usepackage{xspace}
\usepackage{times}
\usepackage{latexsym}
\usepackage{multirow}
\usepackage{algorithm,algorithmicx,algpseudocode}
\usepackage{url}
\usepackage{xcolor}
\usepackage[font={footnotesize}]{caption}
\usepackage{times}  
\usepackage{helvet} 
\usepackage{courier}  
\usepackage{booktabs}       
\usepackage{amsfonts}       
\usepackage{colortbl}
\usepackage{pgfplots}
\pgfplotsset{compat = 1.7}

\DeclareMathOperator*{\argmax}{argmax}
\DeclareUnicodeCharacter{0307}{}

\title{{SimCLS}: A Simple Framework for \\Contrastive Learning of Abstractive Summarization}

\author{Yixin Liu \\
  Carnegie Mellon University \\
  \texttt{yixinl2@cs.cmu.edu} \\\And
  Pengfei Liu \thanks{\ \  Corresponding author.}\\
  Carnegie Mellon University \\
  \texttt{pliu3@cs.cmu.edu} \\}

\date{}

\begin{document}
\maketitle
\begin{abstract}

In this paper, we present a conceptually simple while empirically powerful framework for abstractive summarization,  \textsc{SimCLS}, which can bridge the gap between the \textit{learning objective} and \textit{evaluation metrics} resulting from the currently dominated sequence-to-sequence learning framework by \textbf{formulating text generation as a reference-free evaluation problem} (i.e., quality estimation) assisted by \textit{contrastive learning}.
Experimental results show that, with minor modification over existing top-scoring systems, SimCLS can improve the performance of existing top-performing models by a large margin.
Particularly,  2.51 absolute improvement against BART~\citep{lewis-etal-2020-bart} and 2.50 over PEGASUS~\citep{zhang2020pegasus} w.r.t ROUGE-1 on the CNN/DailyMail dataset, driving the state-of-the-art performance to a new level.
We have open-sourced our codes and results: \url{https://github.com/yixinL7/SimCLS}.
Results of our proposed models have been deployed into \textsc{ExplainaBoard} \cite{liu2021explainaboard} platform, which allows researchers to understand our systems in a more fine-grained way.
\end{abstract}

\section{Introduction}

\begin{figure}[t!]
    \centering
    \includegraphics[width=1\linewidth]{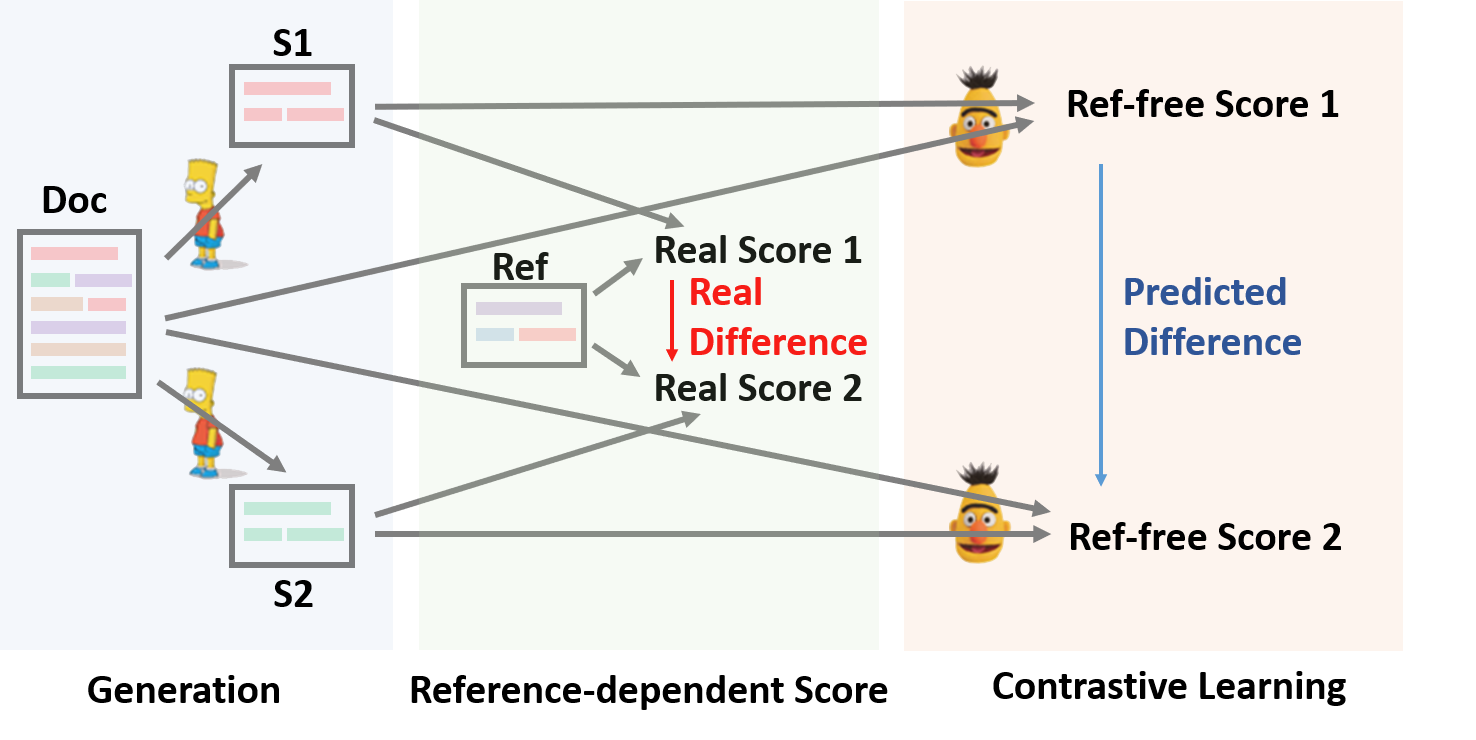}
    \caption{SimCLS framework for two-stage abstractive summarization, where $\text{Doc}$, $\text{S}$, $\text{Ref}$ represent the document, generated summary and reference respectively.  At the first stage, a Seq2Seq generator (BART) is used to generate candidate summaries. At the second stage, a scoring model (RoBERTa) is used to predict the performance of the candidate summaries based on the source document. The scoring model is trained with contrastive learning, where the training examples are provided by the Seq2Seq model.}
    \label{fig:intro}
\end{figure}

Sequence-to-sequence (Seq2Seq) neural models~\citep{10.5555/2969033.2969173} have been widely used for language generation tasks, such as abstractive summarization~\citep{nallapati-etal-2016-abstractive} and neural machine translation~\citep{DBLP:journals/corr/WuSCLNMKCGMKSJL16}. 
While abstractive models~\citep{lewis-etal-2020-bart, zhang2020pegasus} have shown promising potentials in the summarization task, they share the widely acknowledged challenges of Seq2Seq model training. 
Specifically, Seq2Seq models are usually trained under the framework of Maximum Likelihood Estimation (MLE) and in practice they are commonly trained with the \textit{teacher-forcing}~\citep{10.1162/neco.1989.1.2.270} algorithm. 
This introduces a gap between the \textit{objective function} and the \textit{evaluation metrics}, as the objective function is based on local, token-level predictions while the evaluation metrics (e.g. ROUGE~\citep{lin-2004-rouge}) would compare the holistic similarity between the gold references and system outputs.
Furthermore, during the test stage the model needs to generate outputs autoregressivelly, which means the errors made in the previous steps will accumulate. 
This gap between the \textit{training} and \textit{test} has been referred to as the \textit{exposure bias} in the previous work~\citep{10.5555/2969239.2969370, DBLP:journals/corr/RanzatoCAZ15}.

A main line of approaches~\citep{paulus2018a, li-etal-2019-deep} proposes to use the paradigm of Reinforcement Learning (RL) to mitigate the aforementioned gaps.
While RL training makes it possible to train the model with rewards based on global predictions and closely related to the evaluation metrics, it introduces the common challenges of deep RL. 
Specifically, RL-based training suffers from the noise gradient estimation~\citep{greensmith2004variance} problem, which often makes the training unstable and sensitive to hyper-parameters.
Minimum risk training, as an alternative, has also been used in the language generation tasks~\citep{shen-etal-2016-minimum, wieting-etal-2019-beyond}. However, the accuracy of the estimated loss is restricted by the number of sampled outputs.
Other methods~\citep{wiseman-rush-2016-sequence, NIPS2016_2f885d0f, edunov-etal-2018-classical} aim to extend the framework of MLE to incorporate sentence-level scores into the objective functions. 
While these methods can mitigate the limitations of MLE training, the relation between the evaluation metrics and the objective functions used in their methods can be indirect and implicit. 

Among this background, in this work we generalize the paradigm of contrastive learning~\citep{chopra2005learning} to introduce an approach for abstractive summarization which achieves the goal of directly optimizing the model with the corresponding evaluation metrics, thereby mitigating the gaps between training and test stages in MLE training.
While some related work~\citep{lee2021contrastive, pan2021contrastive} have proposed to introduce a contrastive loss as an augmentation of MLE training for conditional text generation tasks, we instead choose to disentangle the functions of contrastive loss and MLE loss by introducing them at different stages in our proposed framework.

Specifically, inspired by the recent work of \citet{zhong-etal-2020-extractive, liu-etal-2021-refsum} on text summarization, we propose to use a two-stage model for abstractive summarization, where a Seq2Seq model is first trained to generate candidate summaries with MLE loss, and then a parameterized evaluation model is trained to rank the generated candidates with contrastive learning.
By optimizing the generation model and evaluation model at separate stages, we are able to train these two modules with supervised learning, bypassing the challenging and intricate optimization process of the RL-based methods.

Our main contribution in this work is to approach metric-oriented training for abstractive summarization by proposing a generate-then-evaluate two-stage framework with contrastive learning, which not only put the state-of-the-art performance on CNN/DailyMail to a new level (2.2 ROUGE-1 improvement against the baseline model), also demonstrates the great potentials of this two-stage framework, calling for future efforts on optimizing Seq2Seq models using methods beyond maximum likelihood estimation.

\section{Contrastive Learning Framework for Abstractive Summarization}

Given a source document $D$ and a reference summary $\hat{S}$, the goal of an abstractive summarization model $f$ is to generate the candidate summary $S = f(D)$ such that it receives the highest score $m = M(S, \hat{S}) $ assigned by an evaluation metric $M$.
In this work, we break down the holistic generation process into two stages which consist of a \textit{generation model} $g$ for generating candidate summaries and a \textit{evaluation model} $h$ for scoring and selecting the best candidate. 
Fig~\ref{fig:intro} illustrates the general framework.

\paragraph{Stage I: Candidate Generation}
The generation model $g(\cdot)$ is a Seq2Seq model trained to maximize the likelihood of reference summary $\hat{S}$ given the source document $D$.
The pre-trained $g(\cdot)$ is then used to produce multiple candidate summaries $S_1, \cdots, S_n$ with a sampling strategy such as Beam Search, where $n$ is the number of sampled candidates.

\noindent\textbf{Stage II: Reference-free Evaluation}
The high-level idea is that a better candidate summary $S_i$ should obtain a higher quality score w.r.t the source document $D$.
We approach the above idea by contrastive learning and define an \textit{evaluation function}
$h(\cdot)$ that aims to assign different scores $r_1, \cdots, r_n$ to the generated candidates solely based on the similarity between the source document and the candidate $S_i$, i.e., $r_i = h(S_i, D)$. 
The final output summary $S$ is the candidate with the highest score:
\begin{align}
    S = \argmax_{S_i}h(S_i, D).
\end{align}
Here, we instantiate $h(\cdot)$ as a large pre-trained self-attention model, RoBERTa~\citep{DBLP:journals/corr/abs-1907-11692}. 
It is used to encode $S_i$ and $D$ separately, and the cosine similarity between the encoding of the first tokens is used as the similarity score $r_i$. 

\paragraph{Contrastive Training}
Instead of explicitly constructing a positive or negative example as most existing work with contrastive learning have adopted \citep{chen2020simple,wu-etal-2020-unsupervised}, here the ``\textit{contrastiveness}'' is reflect in the diverse qualities of naturally generated summaries evaluated by a parameterized model $h(\cdot)$.
Specifically, we introduce a ranking loss to $h(\cdot)$:
\begin{equation}
\small
\begin{split}
      L &= \sum_i \max(0, h(D, \tilde{S}_i) - h(D, \hat{S})) \\
     & + \sum_i \sum_{j > i} \max(0, h(D, \tilde{S}_j) - h(D, \tilde{S}_i) + \lambda_{ij}),   \\
\end{split}
\end{equation}
where $\tilde{S}_1, \cdots, \tilde{S}_n$ is descendingly sorted by $M(\tilde{S}_i, \hat{S})$.
Here, $\lambda_{ij} = (j - i) * \lambda$ is the corresponding margin that we defined following \citet{zhong-etal-2020-extractive}, and $\lambda$ is a hyper-parameter.\footnote{As it is insensitive, we fix it to 0.01 in our experiments.}
$M$ can be any automated evaluation metrics or human judgments and here we use ROUGE~\citep{lin-2004-rouge}.

\section{Experiments}

\subsection{Datasets}
We use two datasets for our experiments. 
The dataset statistics are listed in Appendix \ref{app:data}.

\noindent\texttt{CNNDM} CNN/DailyMail\footnote{\url{https://cs.nyu.edu/~kcho/DMQA/}}~\citep{10.5555/2969239.2969428,nallapati-etal-2016-abstractive} dataset is a large scale news articles dataset.

\noindent\texttt{XSum}
XSum\footnote{\url{https://github.com/EdinburghNLP/XSum}}~\citep{narayan-etal-2018-dont} dataset is a highly abstractive dataset containing online articles from the British Broadcasting Corporation (BBC).

\subsection{Evaluation Metrics}

We use ROUGE-1/2/L (R-1/2/L) as the main evaluation metrics for our experiments. 
We also evaluate our model on the recently developed semantic similarity metrics, namely, BERTScore~\citep{DBLP:conf/iclr/ZhangKWWA20} and MoverScore~\citep{zhao-etal-2019-moverscore}.

\subsection{Base Systems}

As the generation model and the evaluation model in our two-stage framework are trained separately, we use pre-trained state-of-the-art abstractive summarization systems as our generation model. 
Specifically, we use \textbf{BART}~\citep{lewis-etal-2020-bart} and \textbf{Pegasus}~\citep{zhang2020pegasus} as they are popular and have been comprehensively evaluated.

\subsection{Training Details}
For baseline systems, we use the checkpoints provided by the \textit{Transformers}\footnote{\url{https://github.com/huggingface/transformers}}~\citep{wolf-etal-2020-transformers} library. 
We use diverse beam search~\citep{DBLP:journals/corr/VijayakumarCSSL16} as the sampling strategy to generate candidate summaries.
We use 16 groups for diversity sampling, which results in 16 candidates.
To train the evaluation model, we use Adam optimizer~\citep{DBLP:journals/corr/KingmaB14} with learning rate scheduling.
The model performance on the validation set is used to select the checkpoint.
More details are described in Appendix \ref{app:exp}. 

\subsection{Results on CNNDM dataset}
\begin{table}[t!]
\small
\centering
\addtolength{\tabcolsep}{-1pt}  
\begin{tabular}{lccccc}
\toprule
\textbf{System} & \textbf{R-1} & \textbf{R-2} & \textbf{R-L} & \textbf{BS} & \textbf{MS} \\
\midrule
 BART* & 44.16 & 21.28 & 40.90 & - & -\\
 Pegasus* & 44.17 & 21.47 & 41.11 & - & - \\
 Prophet* & 44.20 & 21.17 & 41.30 & - & - \\
 GSum* & 45.94 & \textbf{22.32} & 42.48 & - & -\\
\midrule
 Origin & 44.39 & 21.21 & 41.28  & 64.67 & 58.67  \\
 Min & 33.17 & 11.67 & 30.77 & 58.09 & 55.75 \\
 Max & 54.36 & 28.73 & 50.77 & 70.77 & 61.67 \\
 Random & 43.98 & 20.06 & 40.94 & 64.65 & 58.60 \\
 \midrule
 SimCLS & $\textbf{46.67}^{\dag}$ & $22.15^{\dag}$ & $\textbf{43.54}^{\dag}$ & $\textbf{66.14}^\dag$ & $\textbf{59.31}^\dag$ \\
\bottomrule
\end{tabular}
\addtolength{\tabcolsep}{+1pt} 
\caption{\label{tab:cnndm} Results on \texttt{CNNDM}.
\textbf{BS} denotes BERTScore, \textbf{MS} denotes MoverScore. 
\textbf{Origin} denotes the original performance of the baseline model.
\textbf{Min}, \textbf{Max}, \textbf{Random} are the oracles that select candidates based on their ROUGE scores.
\dag: significantly better than the baseline model (Origin) ($p < 0.01$).
*: results reported in the original papers.}
\end{table}

\begin{figure}
    \raggedright
    \begin{tikzpicture}[scale=0.8]
    \begin{axis}[
    xlabel={Number of Test Candidates},
    xtick={0,4,8,12,16},
    xmin=0,
    ylabel={ROUGE-1},
    label style={font=\large},
    tick label style={font=\large},
    legend entries={SimCLS, Origin.},
    mark size=1.0pt,
    ymajorgrids=true,
    grid style=dashed,
    legend style={font=\large,line width=.5pt,mark size=1.5pt,
            at={(0.6,0.6)},
            anchor=south west,
            /tikz/every even column/.append style={column sep=0.5em}},
            smooth,
    ]
    \addplot [cyan,mark=*,opacity=0.6, line width=2.5pt] table [x=Num, y=ROUGE-1] {./num.txt};
    \addplot [gray,mark=*,opacity=0.6, line width=2.5pt] table [x=Num, y=ROUGE-1] {./baseline.txt};
    \end{axis}
    \end{tikzpicture}
    \caption{Test performance with different numbers of candidate summaries on \texttt{CNNDM}. \textbf{Origin} denotes the original performance of the baseline model.}
    \label{fig:num}
\end{figure}
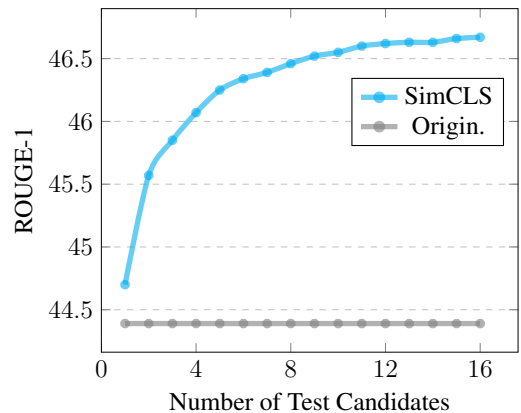

The results on \texttt{CNNDM} dataset are shown in Tab.~\ref{tab:cnndm}.
We use the pretrained BART\footnote{`facebook/bart-large-cnn'} as the base generation model (\textbf{Origin}). We use \textbf{BART}, \textbf{Pegasus}, \textbf{GSum}~\citep{dou-etal-2021-gsum} and \textbf{ProphetNet}~\citep{qi-etal-2020-prophetnet} for comparison.
Notably, the Max oracle which always selects the best candidate has much better performance than the original outputs, suggesting that using a diverse sampling strategy can further exploit the potential power of the pre-trained abstractive system. 
Apart from ROUGE, we also present the evaluation results on semantic similarity metrics.
Our method is able to outperform the baseline model on all metrics, demonstrating its improvement is beyond exploiting the potential artifacts of ROUGE.
While the scale of improvement is harder to interpret with these metrics, we note that the improvement is able to pass the significance test. 

\begin{table*}[t!]
    \scriptsize
    \centering
    \extrarowheight=\aboverulesep
    \addtolength{\extrarowheight}{\belowrulesep}
    \aboverulesep=1pt
    \belowrulesep=1pt
    \begin{tabular}{@{}c  p{0.32\textwidth} p{0.54\textwidth}}
     \toprule
   \multicolumn{1}{c}{ \bf System} & \multicolumn{1}{c}{ \bf Summary} &  \multicolumn{1}{c}{ \bf Article} \\\midrule
     \multicolumn{1}{c}{\bf Ref.} &  chris ramsey says he has no problem shaking hands with john terry . queens park rangers host chelsea in the premier league on sunday . terry was once banned and fined for racist comments at loftus road . rio ferdinand , brother of anton , will not be fit to play against chelsea . &  \multirow{3}{*}{\parbox[height=1.5\textwidth]{0.54\textwidth} {{\bf queens park rangers manager chris ramsey has revealed he will have no problem shaking john terry's hand in light of the racist comments the former england captain directed at former rs defender anton ferdinand four years ago .} {\bf \textit{terry , who will line up against ramsey's side , was banned for four games and fined \# 220,000 for the remarks made in october 2011 during chelsea's 1-0 defeat at loftus road .}} but ramsey , the premier league's only black manager , thinks the issue has been dealt with . ... ` i don't know what his feelings are towards me . as long as there wasn't anything on the field that was unprofessional by him , i would shake his hand . . {\bf queens park rangers manager chris ramsey speaks to the media on friday ahead of the chelsea match .} chelsea captain john terry controls the ball during last weekend's premier league match against stoke . ramsey arrives for friday's pre-match press conference as qpr prepare to host chelsea at loftus road . ` the whole episode for british society sat uncomfortably . it's not something we want to highlight in football . it happened and it's being dealt with . we have to move on . and hopefully everyone has learned something from it . ' . \textit{ramsey revealed that rio ferdinand , who labelled terry an idiot for the abuse aimed at his brother , won't be fit in time for a reunion with the chelsea skipper this weekend .} but the 52-year-old suspects his player's one-time england colleague will be on the receiving end of a hostile welcome from the home fans on his return the scene of the unsavoury incident . ... ferdinand and terry argue during qpr's 1-0 victory against chelsea at loftus road in october 2011 . {\bf rio ferdinand , brother of anton , will not be fit for sunday's match against chelsea .}}}
 \\\cmidrule{1-2}
    \multicolumn{1}{c}{\cellcolor{gray!25}\bf SimCLS} 
     & \cellcolor{gray!25} queens park rangers host chelsea in the premier league on sunday . qpr boss chris ramsey says he will have no problem shaking john terry's hand . terry was banned for four games and fined \# 220,000 for racist comments . rio ferdinand , brother of anton , will not be fit for the match at loftus road . & 
 \\\cmidrule{1-2}
 \multicolumn{1}{c}{\bf Origin.} 
     &  john terry was banned for four games and fined \# 220,000 for the remarks made in october 2011 during chelsea's 1-0 defeat at loftus road . terry will line up against chris ramsey's side on sunday . rio ferdinand , who labelled terry an idiot for the abuse aimed at his brother , won't be fit in time for a reunion with the chelsea skipper this weekend . & 
 \\
  \bottomrule
\end{tabular}
\caption{Sentence alignments between source articles and summaries on \texttt{CNNDM} dataset. The aligned sentences for reference and our summaries are {\bf bolded} (they are the same in this example). The aligned sentences for baseline summaries are {\it italicized}. \textbf{Origin} denotes the original performance of the baseline model.}
\label{tab:example}
\end{table*}

With the constraints of computation power, we try to use as many candidates as possible for the evaluation model training. 
However, we also notice that our method is robust to the specific number of candidates, as during test we found that our model is still able to outperform the baseline model with fewer candidates, which is illustrated in Fig.~\ref{fig:num}.

\begin{table}[t!]
\small
\centering
\begin{tabular}{lcccc}
\toprule
\textbf{Level} & \textbf{System} & \textbf{Precision} & \textbf{Recall} & \textbf{F-Score}\\
\midrule
 \multirow{2}{*}{\parbox[]{0.08\textwidth}{Entity}} & Origin & 40.70 & 59.13 & 48.22 \\
 & SimCLS & \textbf{43.36} & \textbf{59.79} & \textbf{50.27}\\
\midrule
  \multirow{2}{*}{\parbox[]{0.08\textwidth}{Sentence}} & Origin & 38.11 & 38.65 & 37.18 \\
 & SimCLS & \textbf{42.58} & \textbf{40.22} & \textbf{40.12}\\
\bottomrule
\end{tabular}
\caption{\label{tab:entities} Performance analysis on \texttt{CNNDM} dataset. \textbf{Origin} denotes the original performance of the baseline model.}
\end{table}

\subsection{Fine-grained Analysis}
To demonstrate that our method is able to make meaningful improvement w.r.t the summary quality, here we compare our method with the baseline model at different semantic levels on \texttt{CNNDM}.

\subsubsection{Entity-level}
Inspired by the work of \citet{gekhman-etal-2020-kobe} and \citet{jain-etal-2020-scirex}, we compare the model performance w.r.t the \textit{salient entities}, which are entities in source documents that appear in the reference summaries.
Specifically, (1) we extract the entities from the source documents,\footnote{We use a pre-trained NER model provided by spaCy to extract the entities: \url{https://spacy.io/}} (2) select the \textit{salient entities} based on the entities in reference summaries, (3) compare the \textit{salient entities} with entities in candidate summaries.
Results in Tab.~\ref{tab:entities} demonstrate that our method can better capture the important semantic information of the source documents.

\subsubsection{Sentence-level} 
\paragraph{Sentence Alignments}
Here we investigate if our method makes sentence-level differences compared to the baseline model.
Specifically, (1) we match each sentence in the summaries to a sentence in the source documents based on their similarity (indicated by ROUGE scores),\footnote{Notably, this matching approach formulates an extractive oracle when reference summaries are used for matching, which achieves 54.54/30.73/50.35 ROUGE-1/2/L scores.} (2) compute the sentence-level similarity between the reference and system-generated summaries based on the overlaps of their matched sentences in the source documents.
The results in Tab.~\ref{tab:entities} demonstrate that the generated summaries of our method is more similar to the reference summaries at the sentence level.

\pgfplotsset{every tick label/.append style={font=\small}}
\pgfplotsset{every axis label/.append style={font=\small}}

\paragraph{Positional Bias} In Tab.~\ref{tab:example}, we present a case study of the sentence alignment. 
We use the same matching approach to map the summary sentences to the sentences in source articles.
In this example, the output of our method focuses on the same sentences as the reference summary does, while the baseline summary focuses on some different sentences. 

\begin{figure}
    \centering
    \begin{tikzpicture}
\raggedleft
  \begin{axis}[ybar=1.0pt,
    height=0.3\textwidth,
    width=0.5\textwidth,
    bar width=3.5pt,
    ylabel shift=-2pt,
    enlarge y limits={upper,value=0.15},
    axis lines*=left,
    legend style={nodes={scale=0.8, transform shape}, at={(0.5,-0.3)},anchor=north,legend columns=-1
    /tikz/every even column/.append style={column sep=0.05\textwidth}},
    ylabel={Ratio $(\%)$},
    xlabel={Relative Position$(\%)$},
    xticklabels={$10$, $20$, $30$, $40$, $50$, $60$, $70$, $80$, $90$, $100$},
    xtick={1,...,10},
    xmajorgrids=true,
    ymajorgrids=true,
    zmajorgrids=true,
    grid style=dashed,
    xticklabel style={
        inner sep=0.4pt,
    },
    ]
    \addplot [draw=black!100,fill=cyan!20] table[x index=0,y index=1]{plot.txt};
    \addplot [draw=black!100,fill=red!20] table[x index=0,y index=2]{plot.txt};
    \addplot [draw=black!100,fill=green!20] table[x index=0,y index=3]{plot.txt};
    \legend{Ref., SimCLS, Origin.}
  \end{axis}
\end{tikzpicture}
    \caption{Positional Bias. X-asis: the relative position of the matched sentence in source documents. Y-axis: the ratio of the matched sentences. For fair comparison, articles are first truncated to the generator's maximum input length. \textbf{Origin} denotes the original performance of the baseline model.}
    \label{fig:bias}
\end{figure}
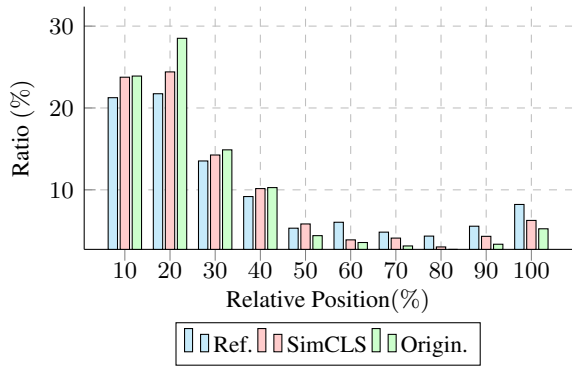

Interestingly, the reference summary focuses on the very last sentence in the article, and our method can follow this pattern.
Upon examining this pattern, we notice a positional bias of abstractive models when handling long source articles (more than 30 sentences).
Fig.~\ref{fig:bias} shows that the baseline summaries are more likely to focus on the head sentences compared to the references, which may result from the autoregressive generation process of the Seq2Seq models.
Our method is able to mitigate this bias, as the candidate sampling process (diverse beam search) generates candidates different from the original outputs, and our evaluation model can assess the holistic quality of the candidates.

\subsection{Results on \texttt{XSum} dataset}

To evaluate our method's performance beyond \texttt{CNNDM} dataset, we also test our method on \texttt{XSum} dataset, and the results are shown in Tab.~\ref{tab:other}.
Here, we use Pegasus\footnote{`google/pegasus-xsum'} as the base system since it achieves better performance than BART on \texttt{XSum}.
We follow the same sampling strategy to generate the training data.
However, as this strategy generally results in lower ROUGE-2 score on \texttt{XSum} dataset, we use a different strategy to generate the validation and test data (4 candidates generated by 4 diverse groups).
Our method is still able to outperform the baseline, but with a smaller margin compared to \texttt{CNNDM}.
Summaries in \texttt{XSum} are shorter (one-sentence) and more abstractive, which restricts the semantic diversity of candidates and makes it harder to make meaningful improvement.
\begin{table}[t!]
\small
\centering
\addtolength{\tabcolsep}{-1pt} 
\begin{tabular}{lccccc}
\toprule
\textbf{System} & \textbf{R-1} & \textbf{R-2} & \textbf{R-L}  & \textbf{BS} & \textbf{MS}\\
 \midrule
 BART* & 45.14 & 22.27 & 37.25 & - & -\\
 Pegasus* & 47.21 & 24.56 & 39.25 & - & - \\
 GSum* & 45.40 & 21.89 & 36.67 & - & -\\
\midrule
 Origin & 47.10 & 24.53 & 39.23 & 69.48 & 61.34\\
 Min & 40.97 & 19.18 & 33.68 & 66.01 & 59.58 \\
 Max & 52.45 & 28.28 & 43.36 & 72.56 & 62.98\\
 Random & 46.72 & 23.64 & 38.55 & 69.30 & 61.23\\
\midrule
 SimCLS & $\textbf{47.61}^\dag$ & \textbf{24.57} & $\textbf{39.44}^\dag$ & $\textbf{69.81}^\dag$ & $\textbf{61.48}^\dag$\\
\bottomrule
\end{tabular}
\addtolength{\tabcolsep}{+1pt} 
\vspace{-2mm}
\caption{\label{tab:other} Results on \texttt{XSum} dataset. 
\textbf{BS} denotes BERTScore, \textbf{MS} denotes MoverScore. 
\textbf{Origin} denotes the original performance of the baseline model.
\textbf{Min}, \textbf{Max}, \textbf{Random} are the oracles that select candidates based on their ROUGE scores.
\dag: significantly better than the baseline model (Origin) ($p < 0.05$). 
*: results reported in the original papers.}
\end{table}

\section{Conclusion}

In this work, we present a contrastive summarization framework that aims to optimize the quality of generated summaries at summary-level, which mitigates the discrepancy between the training and test stages in the MLE framework.
Apart from the significant improvement over the baseline model on \texttt{CNNDM} dataset, we present a comprehensive evaluation at different semantic levels, explaining the sources of the improvement made by our method.
Notably, our experimental results also indicate that the existing abstractive systems have the potential of generating candidate summaries much better than the original outputs. 
Therefore, our work opens up the possibility for future directions including (1) extending this two-stage strategy to other datasets for abstractive models; (2) improving the training algorithms for abstractive models towards a more holistic optimization process.

\section*{Acknowledgements}

We thank Professor Graham Neubig and anonymous reviewers for valuable feedback and helpful suggestions.
This work was supported in part by a grant under the Northrop Grumman SOTERIA project and the Air Force Research Laboratory under agreement number FA8750-19-2-0200. The U.S. Government
is authorized to reproduce and distribute reprints for Governmental
purposes notwithstanding any copyright notation thereon. The views and
conclusions contained herein are those of the authors and should not be
interpreted as necessarily representing the official policies or
endorsements, either expressed or implied, of the Air Force Research
Laboratory or the U.S. Government.

\bibliographystyle{acl_natbib}

\bibliography{anthology,acl2021}

\begin{thebibliography}{37}
\expandafter\ifx\csname natexlab\endcsname\relax\def\natexlab#1{#1}\fi

\bibitem[{Bengio et~al.(2015)Bengio, Vinyals, Jaitly, and
  Shazeer}]{10.5555/2969239.2969370}
Samy Bengio, Oriol Vinyals, Navdeep Jaitly, and Noam Shazeer. 2015.
\newblock Scheduled sampling for sequence prediction with recurrent neural
  networks.
\newblock In \emph{Proceedings of the 28th International Conference on Neural
  Information Processing Systems - Volume 1}, NIPS'15, page 1171–1179,
  Cambridge, MA, USA. MIT Press.

\bibitem[{Chen et~al.(2020)Chen, Kornblith, Norouzi, and
  Hinton}]{chen2020simple}
Ting Chen, Simon Kornblith, Mohammad Norouzi, and Geoffrey Hinton. 2020.
\newblock A simple framework for contrastive learning of visual
  representations.
\newblock In \emph{International conference on machine learning}, pages
  1597--1607. PMLR.

\bibitem[{Chopra et~al.(2005)Chopra, Hadsell, and LeCun}]{chopra2005learning}
Sumit Chopra, Raia Hadsell, and Yann LeCun. 2005.
\newblock Learning a similarity metric discriminatively, with application to
  face verification.
\newblock In \emph{2005 IEEE Computer Society Conference on Computer Vision and
  Pattern Recognition (CVPR'05)}, volume~1, pages 539--546. IEEE.

\bibitem[{Dou et~al.(2021)Dou, Liu, Hayashi, Jiang, and
  Neubig}]{dou-etal-2021-gsum}
Zi-Yi Dou, Pengfei Liu, Hiroaki Hayashi, Zhengbao Jiang, and Graham Neubig.
  2021.
\newblock \href {https://www.aclweb.org/anthology/2021.naacl-main.384} {{GS}um:
  A general framework for guided neural abstractive summarization}.
\newblock In \emph{Proceedings of the 2021 Conference of the North American
  Chapter of the Association for Computational Linguistics: Human Language
  Technologies}, pages 4830--4842, Online. Association for Computational
  Linguistics.

\bibitem[{Edunov et~al.(2018)Edunov, Ott, Auli, Grangier, and
  Ranzato}]{edunov-etal-2018-classical}
Sergey Edunov, Myle Ott, Michael Auli, David Grangier, and Marc{'}Aurelio
  Ranzato. 2018.
\newblock \href {https://doi.org/10.18653/v1/N18-1033} {Classical structured
  prediction losses for sequence to sequence learning}.
\newblock In \emph{Proceedings of the 2018 Conference of the North {A}merican
  Chapter of the Association for Computational Linguistics: Human Language
  Technologies, Volume 1 (Long Papers)}, pages 355--364, New Orleans,
  Louisiana. Association for Computational Linguistics.

\bibitem[{Gekhman et~al.(2020)Gekhman, Aharoni, Beryozkin, Freitag, and
  Macherey}]{gekhman-etal-2020-kobe}
Zorik Gekhman, Roee Aharoni, Genady Beryozkin, Markus Freitag, and Wolfgang
  Macherey. 2020.
\newblock \href {https://doi.org/10.18653/v1/2020.findings-emnlp.287}
  {{K}o{BE}: Knowledge-based machine translation evaluation}.
\newblock In \emph{Findings of the Association for Computational Linguistics:
  EMNLP 2020}, pages 3200--3207, Online. Association for Computational
  Linguistics.

\bibitem[{Greensmith et~al.(2004)Greensmith, Bartlett, and
  Baxter}]{greensmith2004variance}
Evan Greensmith, Peter~L Bartlett, and Jonathan Baxter. 2004.
\newblock Variance reduction techniques for gradient estimates in reinforcement
  learning.
\newblock \emph{Journal of Machine Learning Research}, 5(9).

\bibitem[{Hermann et~al.(2015)Hermann, Ko\v{c}isk\'{y}, Grefenstette, Espeholt,
  Kay, Suleyman, and Blunsom}]{10.5555/2969239.2969428}
Karl~Moritz Hermann, Tom\'{a}\v{s} Ko\v{c}isk\'{y}, Edward Grefenstette, Lasse
  Espeholt, Will Kay, Mustafa Suleyman, and Phil Blunsom. 2015.
\newblock Teaching machines to read and comprehend.
\newblock In \emph{Proceedings of the 28th International Conference on Neural
  Information Processing Systems - Volume 1}, NIPS'15, page 1693–1701,
  Cambridge, MA, USA. MIT Press.

\bibitem[{Jain et~al.(2020)Jain, van Zuylen, Hajishirzi, and
  Beltagy}]{jain-etal-2020-scirex}
Sarthak Jain, Madeleine van Zuylen, Hannaneh Hajishirzi, and Iz~Beltagy. 2020.
\newblock \href {https://doi.org/10.18653/v1/2020.acl-main.670} {{S}ci{REX}:
  {A} challenge dataset for document-level information extraction}.
\newblock In \emph{Proceedings of the 58th Annual Meeting of the Association
  for Computational Linguistics}, pages 7506--7516, Online. Association for
  Computational Linguistics.

\bibitem[{Kingma and Ba(2015)}]{DBLP:journals/corr/KingmaB14}
Diederik~P. Kingma and Jimmy Ba. 2015.
\newblock \href {http://arxiv.org/abs/1412.6980} {Adam: {A} method for
  stochastic optimization}.
\newblock In \emph{3rd International Conference on Learning Representations,
  {ICLR} 2015, San Diego, CA, USA, May 7-9, 2015, Conference Track
  Proceedings}.

\bibitem[{Lee et~al.(2021)Lee, Lee, and Hwang}]{lee2021contrastive}
Seanie Lee, Dong~Bok Lee, and Sung~Ju Hwang. 2021.
\newblock \href {https://openreview.net/forum?id=Wga_hrCa3P3} {Contrastive
  learning with adversarial perturbations for conditional text generation}.
\newblock In \emph{International Conference on Learning Representations}.

\bibitem[{Lewis et~al.(2020)Lewis, Liu, Goyal, Ghazvininejad, Mohamed, Levy,
  Stoyanov, and Zettlemoyer}]{lewis-etal-2020-bart}
Mike Lewis, Yinhan Liu, Naman Goyal, Marjan Ghazvininejad, Abdelrahman Mohamed,
  Omer Levy, Veselin Stoyanov, and Luke Zettlemoyer. 2020.
\newblock \href {https://doi.org/10.18653/v1/2020.acl-main.703} {{BART}:
  Denoising sequence-to-sequence pre-training for natural language generation,
  translation, and comprehension}.
\newblock In \emph{Proceedings of the 58th Annual Meeting of the Association
  for Computational Linguistics}, pages 7871--7880, Online. Association for
  Computational Linguistics.

\bibitem[{Li et~al.(2019)Li, Lei, Qin, and Wang}]{li-etal-2019-deep}
Siyao Li, Deren Lei, Pengda Qin, and William~Yang Wang. 2019.
\newblock \href {https://doi.org/10.18653/v1/D19-1623} {Deep reinforcement
  learning with distributional semantic rewards for abstractive summarization}.
\newblock In \emph{Proceedings of the 2019 Conference on Empirical Methods in
  Natural Language Processing and the 9th International Joint Conference on
  Natural Language Processing (EMNLP-IJCNLP)}, pages 6038--6044, Hong Kong,
  China. Association for Computational Linguistics.

\bibitem[{Lin(2004)}]{lin-2004-rouge}
Chin-Yew Lin. 2004.
\newblock \href {https://www.aclweb.org/anthology/W04-1013} {{ROUGE}: A package
  for automatic evaluation of summaries}.
\newblock In \emph{Text Summarization Branches Out}, pages 74--81, Barcelona,
  Spain. Association for Computational Linguistics.

\bibitem[{Liu et~al.(2021{\natexlab{a}})Liu, Fu, Xiao, Yuan, Chang, Dai, Liu,
  Ye, and Neubig}]{liu2021explainaboard}
Pengfei Liu, Jinlan Fu, Yang Xiao, Weizhe Yuan, Shuaicheng Chang, Junqi Dai,
  Yixin Liu, Zihuiwen Ye, and Graham Neubig. 2021{\natexlab{a}}.
\newblock Explainaboard: An explainable leaderboard for nlp.
\newblock \emph{arXiv preprint arXiv:2104.06387}.

\bibitem[{Liu et~al.(2019)Liu, Ott, Goyal, Du, Joshi, Chen, Levy, Lewis,
  Zettlemoyer, and Stoyanov}]{DBLP:journals/corr/abs-1907-11692}
Yinhan Liu, Myle Ott, Naman Goyal, Jingfei Du, Mandar Joshi, Danqi Chen, Omer
  Levy, Mike Lewis, Luke Zettlemoyer, and Veselin Stoyanov. 2019.
\newblock \href {http://arxiv.org/abs/1907.11692} {Roberta: {A} robustly
  optimized {BERT} pretraining approach}.
\newblock \emph{CoRR}, abs/1907.11692.

\bibitem[{Liu et~al.(2021{\natexlab{b}})Liu, Dou, and
  Liu}]{liu-etal-2021-refsum}
Yixin Liu, Zi-Yi Dou, and Pengfei Liu. 2021{\natexlab{b}}.
\newblock \href {https://www.aclweb.org/anthology/2021.naacl-main.113}
  {{R}ef{S}um: Refactoring neural summarization}.
\newblock In \emph{Proceedings of the 2021 Conference of the North American
  Chapter of the Association for Computational Linguistics: Human Language
  Technologies}, pages 1437--1448, Online. Association for Computational
  Linguistics.

\bibitem[{Nallapati et~al.(2016)Nallapati, Zhou, dos Santos, Gu̇l{\c{c}}ehre,
  and Xiang}]{nallapati-etal-2016-abstractive}
Ramesh Nallapati, Bowen Zhou, Cicero dos Santos, {\c{C}}a{\u{g}}lar
  Gu̇l{\c{c}}ehre, and Bing Xiang. 2016.
\newblock \href {https://doi.org/10.18653/v1/K16-1028} {Abstractive text
  summarization using sequence-to-sequence {RNN}s and beyond}.
\newblock In \emph{Proceedings of The 20th {SIGNLL} Conference on Computational
  Natural Language Learning}, pages 280--290, Berlin, Germany. Association for
  Computational Linguistics.

\bibitem[{Narayan et~al.(2018)Narayan, Cohen, and
  Lapata}]{narayan-etal-2018-dont}
Shashi Narayan, Shay~B. Cohen, and Mirella Lapata. 2018.
\newblock \href {https://doi.org/10.18653/v1/D18-1206} {Don{'}t give me the
  details, just the summary! topic-aware convolutional neural networks for
  extreme summarization}.
\newblock In \emph{Proceedings of the 2018 Conference on Empirical Methods in
  Natural Language Processing}, pages 1797--1807, Brussels, Belgium.
  Association for Computational Linguistics.

\bibitem[{Norouzi et~al.(2016)Norouzi, Bengio, Chen, Jaitly, Schuster, Wu, and
  Schuurmans}]{NIPS2016_2f885d0f}
Mohammad Norouzi, Samy Bengio, zhifeng Chen, Navdeep Jaitly, Mike Schuster,
  Yonghui Wu, and Dale Schuurmans. 2016.
\newblock \href
  {https://proceedings.neurips.cc/paper/2016/file/2f885d0fbe2e131bfc9d98363e55d1d4-Paper.pdf}
  {Reward augmented maximum likelihood for neural structured prediction}.
\newblock In \emph{Advances in Neural Information Processing Systems},
  volume~29, pages 1723--1731. Curran Associates, Inc.

\bibitem[{Pan et~al.(2021)Pan, Wang, Wu, and Li}]{pan2021contrastive}
Xiao Pan, Mingxuan Wang, Liwei Wu, and Lei Li. 2021.
\newblock \href {http://arxiv.org/abs/2105.09501} {Contrastive learning for
  many-to-many multilingual neural machine translation}.

\bibitem[{Paulus et~al.(2018)Paulus, Xiong, and Socher}]{paulus2018a}
Romain Paulus, Caiming Xiong, and Richard Socher. 2018.
\newblock \href {https://openreview.net/forum?id=HkAClQgA-} {A deep reinforced
  model for abstractive summarization}.
\newblock In \emph{International Conference on Learning Representations}.

\bibitem[{Qi et~al.(2020)Qi, Yan, Gong, Liu, Duan, Chen, Zhang, and
  Zhou}]{qi-etal-2020-prophetnet}
Weizhen Qi, Yu~Yan, Yeyun Gong, Dayiheng Liu, Nan Duan, Jiusheng Chen, Ruofei
  Zhang, and Ming Zhou. 2020.
\newblock \href {https://doi.org/10.18653/v1/2020.findings-emnlp.217}
  {{P}rophet{N}et: Predicting future n-gram for
  sequence-to-{S}equence{P}re-training}.
\newblock In \emph{Findings of the Association for Computational Linguistics:
  EMNLP 2020}, pages 2401--2410, Online. Association for Computational
  Linguistics.

\bibitem[{Ranzato et~al.(2016)Ranzato, Chopra, Auli, and
  Zaremba}]{DBLP:journals/corr/RanzatoCAZ15}
Marc'Aurelio Ranzato, Sumit Chopra, Michael Auli, and Wojciech Zaremba. 2016.
\newblock \href {http://arxiv.org/abs/1511.06732} {Sequence level training with
  recurrent neural networks}.
\newblock In \emph{4th International Conference on Learning Representations,
  {ICLR} 2016, San Juan, Puerto Rico, May 2-4, 2016, Conference Track
  Proceedings}.

\bibitem[{Shen et~al.(2016)Shen, Cheng, He, He, Wu, Sun, and
  Liu}]{shen-etal-2016-minimum}
Shiqi Shen, Yong Cheng, Zhongjun He, Wei He, Hua Wu, Maosong Sun, and Yang Liu.
  2016.
\newblock \href {https://doi.org/10.18653/v1/P16-1159} {Minimum risk training
  for neural machine translation}.
\newblock In \emph{Proceedings of the 54th Annual Meeting of the Association
  for Computational Linguistics (Volume 1: Long Papers)}, pages 1683--1692,
  Berlin, Germany. Association for Computational Linguistics.

\bibitem[{Sutskever et~al.(2014)Sutskever, Vinyals, and
  Le}]{10.5555/2969033.2969173}
Ilya Sutskever, Oriol Vinyals, and Quoc~V. Le. 2014.
\newblock Sequence to sequence learning with neural networks.
\newblock In \emph{Proceedings of the 27th International Conference on Neural
  Information Processing Systems - Volume 2}, NIPS'14, page 3104–3112,
  Cambridge, MA, USA. MIT Press.

\bibitem[{Vijayakumar et~al.(2016)Vijayakumar, Cogswell, Selvaraju, Sun, Lee,
  Crandall, and Batra}]{DBLP:journals/corr/VijayakumarCSSL16}
Ashwin~K. Vijayakumar, Michael Cogswell, Ramprasaath~R. Selvaraju, Qing Sun,
  Stefan Lee, David~J. Crandall, and Dhruv Batra. 2016.
\newblock \href {http://arxiv.org/abs/1610.02424} {Diverse beam search:
  Decoding diverse solutions from neural sequence models}.
\newblock \emph{CoRR}, abs/1610.02424.

\bibitem[{Wieting et~al.(2019)Wieting, Berg-Kirkpatrick, Gimpel, and
  Neubig}]{wieting-etal-2019-beyond}
John Wieting, Taylor Berg-Kirkpatrick, Kevin Gimpel, and Graham Neubig. 2019.
\newblock \href {https://doi.org/10.18653/v1/P19-1427} {Beyond {BLEU}:training
  neural machine translation with semantic similarity}.
\newblock In \emph{Proceedings of the 57th Annual Meeting of the Association
  for Computational Linguistics}, pages 4344--4355, Florence, Italy.
  Association for Computational Linguistics.

\bibitem[{Williams and Zipser(1989)}]{10.1162/neco.1989.1.2.270}
Ronald~J. Williams and David Zipser. 1989.
\newblock \href {https://doi.org/10.1162/neco.1989.1.2.270} {A learning
  algorithm for continually running fully recurrent neural networks}.
\newblock \emph{Neural Comput.}, 1(2):270–280.

\bibitem[{Wiseman and Rush(2016)}]{wiseman-rush-2016-sequence}
Sam Wiseman and Alexander~M. Rush. 2016.
\newblock \href {https://doi.org/10.18653/v1/D16-1137} {Sequence-to-sequence
  learning as beam-search optimization}.
\newblock In \emph{Proceedings of the 2016 Conference on Empirical Methods in
  Natural Language Processing}, pages 1296--1306, Austin, Texas. Association
  for Computational Linguistics.

\bibitem[{Wolf et~al.(2020)Wolf, Debut, Sanh, Chaumond, Delangue, Moi, Cistac,
  Rault, Louf, Funtowicz, Davison, Shleifer, von Platen, Ma, Jernite, Plu, Xu,
  Le~Scao, Gugger, Drame, Lhoest, and Rush}]{wolf-etal-2020-transformers}
Thomas Wolf, Lysandre Debut, Victor Sanh, Julien Chaumond, Clement Delangue,
  Anthony Moi, Pierric Cistac, Tim Rault, Remi Louf, Morgan Funtowicz, Joe
  Davison, Sam Shleifer, Patrick von Platen, Clara Ma, Yacine Jernite, Julien
  Plu, Canwen Xu, Teven Le~Scao, Sylvain Gugger, Mariama Drame, Quentin Lhoest,
  and Alexander Rush. 2020.
\newblock \href {https://doi.org/10.18653/v1/2020.emnlp-demos.6} {Transformers:
  State-of-the-art natural language processing}.
\newblock In \emph{Proceedings of the 2020 Conference on Empirical Methods in
  Natural Language Processing: System Demonstrations}, pages 38--45, Online.
  Association for Computational Linguistics.

\bibitem[{Wu et~al.(2020)Wu, Ma, Wu, Manyumwa, and
  Ji}]{wu-etal-2020-unsupervised}
Hanlu Wu, Tengfei Ma, Lingfei Wu, Tariro Manyumwa, and Shouling Ji. 2020.
\newblock \href {https://doi.org/10.18653/v1/2020.emnlp-main.294} {Unsupervised
  reference-free summary quality evaluation via contrastive learning}.
\newblock In \emph{Proceedings of the 2020 Conference on Empirical Methods in
  Natural Language Processing (EMNLP)}, pages 3612--3621, Online. Association
  for Computational Linguistics.

\bibitem[{Wu et~al.(2016)Wu, Schuster, Chen, Le, Norouzi, Macherey, Krikun,
  Cao, Gao, Macherey, Klingner, Shah, Johnson, Liu, Kaiser, Gouws, Kato, Kudo,
  Kazawa, Stevens, Kurian, Patil, Wang, Young, Smith, Riesa, Rudnick, Vinyals,
  Corrado, Hughes, and Dean}]{DBLP:journals/corr/WuSCLNMKCGMKSJL16}
Yonghui Wu, Mike Schuster, Zhifeng Chen, Quoc~V. Le, Mohammad Norouzi, Wolfgang
  Macherey, Maxim Krikun, Yuan Cao, Qin Gao, Klaus Macherey, Jeff Klingner,
  Apurva Shah, Melvin Johnson, Xiaobing Liu, Lukasz Kaiser, Stephan Gouws,
  Yoshikiyo Kato, Taku Kudo, Hideto Kazawa, Keith Stevens, George Kurian,
  Nishant Patil, Wei Wang, Cliff Young, Jason Smith, Jason Riesa, Alex Rudnick,
  Oriol Vinyals, Greg Corrado, Macduff Hughes, and Jeffrey Dean. 2016.
\newblock \href {http://arxiv.org/abs/1609.08144} {Google's neural machine
  translation system: Bridging the gap between human and machine translation}.
\newblock \emph{CoRR}, abs/1609.08144.

\bibitem[{Zhang et~al.(2020{\natexlab{a}})Zhang, Zhao, Saleh, and
  Liu}]{zhang2020pegasus}
Jingqing Zhang, Yao Zhao, Mohammad Saleh, and Peter Liu. 2020{\natexlab{a}}.
\newblock Pegasus: Pre-training with extracted gap-sentences for abstractive
  summarization.
\newblock In \emph{International Conference on Machine Learning}, pages
  11328--11339. PMLR.

\bibitem[{Zhang et~al.(2020{\natexlab{b}})Zhang, Kishore, Wu, Weinberger, and
  Artzi}]{DBLP:conf/iclr/ZhangKWWA20}
Tianyi Zhang, Varsha Kishore, Felix Wu, Kilian~Q. Weinberger, and Yoav Artzi.
  2020{\natexlab{b}}.
\newblock \href {https://openreview.net/forum?id=SkeHuCVFDr} {Bertscore:
  Evaluating text generation with {BERT}}.
\newblock In \emph{8th International Conference on Learning Representations,
  {ICLR} 2020, Addis Ababa, Ethiopia, April 26-30, 2020}. OpenReview.net.

\bibitem[{Zhao et~al.(2019)Zhao, Peyrard, Liu, Gao, Meyer, and
  Eger}]{zhao-etal-2019-moverscore}
Wei Zhao, Maxime Peyrard, Fei Liu, Yang Gao, Christian~M. Meyer, and Steffen
  Eger. 2019.
\newblock \href {https://doi.org/10.18653/v1/D19-1053} {{M}over{S}core: Text
  generation evaluating with contextualized embeddings and earth mover
  distance}.
\newblock In \emph{Proceedings of the 2019 Conference on Empirical Methods in
  Natural Language Processing and the 9th International Joint Conference on
  Natural Language Processing (EMNLP-IJCNLP)}, pages 563--578, Hong Kong,
  China. Association for Computational Linguistics.

\bibitem[{Zhong et~al.(2020)Zhong, Liu, Chen, Wang, Qiu, and
  Huang}]{zhong-etal-2020-extractive}
Ming Zhong, Pengfei Liu, Yiran Chen, Danqing Wang, Xipeng Qiu, and Xuanjing
  Huang. 2020.
\newblock \href {https://doi.org/10.18653/v1/2020.acl-main.552} {Extractive
  summarization as text matching}.
\newblock In \emph{Proceedings of the 58th Annual Meeting of the Association
  for Computational Linguistics}, pages 6197--6208, Online. Association for
  Computational Linguistics.

\end{thebibliography}

\appendix
\section{Dataset Statistics}
\label{app:data}
\begin{table}[h]
  \centering
  \small
    \begin{tabular}{@{\extracolsep{1pt}}lccccc}
    \toprule
    \multirow{2}{*}{Datasets} & \multicolumn{3}{c}{\# Num} & \multicolumn{2}{c}{Avg. Len} \\
    \cmidrule{2-4} \cmidrule{5-6}
    & Train & Valid & Test & Doc. & Sum. \\
    \midrule
    CNNDM & 287K & 13K & 11K & 768.6 & 55.7 \\
    XSum & 203K & 11K & 11K & 429.2 & 23.3 \\
    \bottomrule
    \end{tabular}%
  \caption{Datasets Statistics. Len is the length of tokens.}
  \label{tab:data}%
\end{table}%
The source documents and reference summaries are lower-cased. 
Due to the input length limitation, some source documents are truncated during training.

\section{Experiment Details}
\label{app:exp}
\textbf{Candidate Generation} We use diverse beam search to generate the candidate summaries. 
We use the same beam search configuration as the original work except those related to diverse beam search.
In particular, the diversity penalty is set to 1, and we use 16 diversity groups with 16 beams, which results in 16 candidates.

\noindent \textbf{Model} We use the pretrained RoBERTa with `roberta-base' version provided by the \textit{Transformers} library as our evaluation model, which contains 125M parameters.

\noindent \textbf{Optimizer} We use Adam optimizer with learning rate scheduling:
\begin{align}
    lr &= 0.002 \cdot \min(\mathrm{step\_num}^{-0.5}, \\ \nonumber &\mathrm{step\_num}\cdot\mathrm{warmup\_steps}^{-1.5}),
\end{align}
where the $\mathrm{warmup\_steps}$ is 10000.

\noindent \textbf{Training details} The batch size in our experiments is 32. We evaluate the model performance on the validation set at every 1000 steps, using the averaged ROUGE-1/2/L score as the selecting criteria. The training is converged in 5 epochs, which takes around 40 hours on 4 GTX-1080-Ti GPUs on CNN/DailyMail dataset and 20 hours on XSum dataset.

\end{document}